\documentclass{article}

\usepackage[english]{babel}
\usepackage{graphicx}
\usepackage{fancyhdr} 
\usepackage{authblk}
\usepackage[letterpaper,top=2cm,bottom=2cm,left=3cm,right=3cm,marginparwidth=1.75cm]{geometry}

\usepackage{amsmath}
\usepackage{graphicx}
\usepackage[colorlinks=true, allcolors=blue]{hyperref}
\usepackage{multirow}
\usepackage{placeins}
\usepackage{booktabs}
\usepackage{tabularray}
\title{VET-DINO: Learning Anatomical Understanding Through Multi-View Distillation in Veterinary Imaging}
\author{Andre Dourson, Kylie Taylor, Xiaoli Qiao and Michael Fitzke}
\affil{Mars Petcare}
\begin{document}
\maketitle


\begin{abstract}
\label{lab:sec_abstract}
Self-supervised learning has emerged as a powerful paradigm for training deep neural networks, particularly in medical imaging where labeled data is scarce. While current approaches typically rely on synthetic augmentations of single images, we propose VET-DINO, a framework that leverages a unique characteristic of medical imaging: the availability of multiple standardized views from the same study. Using a series of clinical veterinary radiographs from the same patient study, we enable models to learn view-invariant anatomical structures and develop an implied 3D understanding from 2D projections. We demonstrate our approach on a dataset of 5 million veterinary radiographs from 668,000 canine studies. Through extensive experimentation, including view synthesis and downstream task performance, we show that learning from real multi-view pairs leads to superior anatomical understanding compared to purely synthetic augmentations. VET-DINO achieves state-of-the-art performance on various veterinary imaging tasks. Our work establishes a new paradigm for self-supervised learning in medical imaging that leverages domain-specific properties rather than merely adapting natural image techniques.

\end{abstract}

\section{Introduction}
\label{lab:sec_intro}
Deep learning approaches in medical imaging have traditionally relied on adapting techniques developed for natural images \cite{rajpurkar2017chexnet}. 
However, medical imaging possesses unique characteristics that remain under-exploited in current self-supervised learning frameworks. One such characteristic is the availability of multiple standardized views of the same subject, often captured near-simultaneously under controlled protocols - a property that has no direct parallel in natural image datasets.
While recent self-supervised learning approaches like DINO and DINOv2\cite{caron2021emerging}\cite{oquab2024dinov2} have shown remarkable success using augmented views of the same image, they fail to leverage one of the most valuable aspects of medical imaging: the presence of multiple, geometrically-calibrated views capturing the same anatomical structure from different, standardized angles. This is evident in veterinary radiology, where multiple views are routinely captured to enable accurate diagnosis and surgical planning  \cite{ober2006comparison}.
In this work, we introduce VET-DINO, a self-supervised learning framework that leverages multi-view veterinary studies to learn rich anatomical representations. Rather than manipulating single images to simulate variations, we leverage different radiographic views from the same study as natural variations of the same subject. 
This approach enables the model to learn view-invariant anatomical structures and develop an implied understanding of 3D structures from 2D projections. We demonstrate our approach on a large-scale dataset of 5 million veterinary radiographs, comprising of 668,000 unique canine studies.

Our key contributions include:
\begin{enumerate}
    \item A novel adaptation of self-supervised learning that exploits the multi-view nature of medical imaging protocols.
    \item Demonstration that learning from real multi-view pairs leads to improved representation learning compared to traditional synthetic image augmentations on downstream veterinary tasks.
    \item State-of-the-art performance on various downstream veterinary imaging tasks.
    \end{enumerate}

Our results suggest that leveraging domain-specific properties of medical imaging, rather than simply applying natural image techniques, can lead to more effective and efficient learning of anatomical structures. This work represents a significant step toward self-supervised learning approaches that are truly optimized for medical imaging applications.

\section{Related Work}
\label{lab:sec_related_work}
\subsection{Self-supervised Learning in Computer Vision}
Self-supervised learning has revolutionized computer vision by enabling deep neural networks to learn meaningful representations without manual annotations. Early approaches like SimCLR \cite{chen2020simple} and MoCo \cite{he2020momentum} established the effectiveness of contrastive learning by treating augmented versions of the same image as positive pairs while pushing apart representations of different images. More recently, DINO and DINOv2\cite{caron2021emerging}\cite{oquab2024dinov2} demonstrated that self-distillation with no labels can lead to emergent properties, like semantic segmentation and object discovery. These methods typically rely on carefully crafted synthetic augmentations such as random cropping, color jittering, and geometric transformations to create different views of the same instance.

\subsection{Multi-view Learning}
Multi-view learning has evolved significantly in computer vision, progressing from early work in face recognition to recent advances in self-supervised learning and neural representations. FaceNet introduced in \cite{schroff2015facenet} pioneered the use of triplet loss to learn view-invariant face embeddings, demonstrating that models can learn robust features from natural view variations. Recent work has focused on learning view-invariant representations through self-supervised approaches. ViewCLR in \cite{das2023viewclr} introduced a learnable view-generator that enables video representations to generalize across unseen camera viewpoints, achieving state-of-the-art performance on cross-view action recognition benchmarks. Building on the masked autoencoder framework, MV2MAE in \cite{shah2024mv2mae} proposed a cross-view reconstruction approach for synchronized multi-view videos, using cross-attention mechanisms to reconstruct videos across different viewpoints. While these advances have primarily focused on natural images and videos, the principles of multi-view learning are increasingly relevant to medical imaging, where multiple views of the same anatomy are often acquired.

\subsection{Medical Image Analysis}
The medical imaging community has increasingly adopted self-supervised learning to address the persistent challenge of limited labeled data. Recent work has challenged traditional approaches to representation learning in medical imaging. Notably, \cite{perez2024rad} demonstrated that unimodal self-supervised learning can match or exceed the performance of language-supervised models across diverse medical imaging tasks. This finding fundamentally questions the necessity of language supervision for learning general-purpose biomedical image encoders, particularly given the challenges of obtaining paired imaging-text data in healthcare settings. MAIRA-2 \cite{bannur2024maira} further advanced the field by introducing grounded report generation, combining detailed image understanding with precise anatomical localization. These works highlight a shift towards more sophisticated self-supervised approaches that can capture complex anatomical relationships without relying on manual annotations or potentially limiting text supervision.
Earlier approaches, such as MedAug \cite{vu2021medaug} introduced medical image-specific augmentations, while PCRL \cite{zhou2021preservational} incorporated anatomical priors to maintain clinically relevant features during pre-training. However, most of these methods continued to rely on synthetic transformations of single images, potentially limiting the usage of available anatomical information provided by additional views of a single study.

\subsubsection{Multi-view Learning in Medical Image Analysis}
Multi-view learning in medical imaging has seen limited exploration compared to single-view approaches, despite the inherent availability of multiple views in many imaging studies. Notable works, such as  \cite{chen2019med3d}, explored multi-view representations in the context of 3D data. However, these methods primarily focused on volumetric reconstruction or fusion of cross-sectional slices, rather than leveraging natural multi-view pairings.

In \cite{bertrand2019lateral}  the value of lateral X-ray views in addition to frontal posteroanterior (PA) views for chest radiology were investigated. Using the PadChest dataset, they showed that incorporating lateral views increased the AUC for certain conditions and matched PA view performance for others. While their findings suggest potential benefits, they emphasize the need for more sophisticated methods to fully exploit multi-view data.

An attention-based multi-view classifier for chest X-ray was proposed in \cite{wannenmacher2023studyformer}. By combining convolutional neural networks for feature extraction with a Vision Transformer-based attention mechanism, the model effectively leverages multi-view data to outperform traditional single-view models. Their work demonstrates the potential of multi-view learning to improve multi-label classification performance across 41 labels. While these studies demonstrate the potential of multi-view learning in medical imaging, they primarily focus on classification tasks using relatively small datasets and established architectures. Our work, in contrast, leverages a significantly larger dataset of veterinary radiographs to explore self-supervised representation learning using a novel multi-view framework.

\subsection{Veterinary Imaging}
The application of artificial intelligence in veterinary imaging has gained increasing attention, though significant challenges remain. Recent comprehensive reviews (\cite{hennessey2022artificial},\cite{wilson2022role}, \cite{burti2024artificial}) highlight both the potential and limitations of AI in veterinary diagnostic imaging. These works identify key challenges including the scarcity of large, annotated datasets, the diversity of species and anatomical variations, and the need for standardized imaging protocols. Burti et al. \cite{burti2024artificial} particularly emphasize the importance of developing veterinary-specific approaches rather than directly applying human medical models, given the unique challenges of veterinary imaging. 

\cite{fitzke2021rapidread} introduced a system that leverages NLP-derived labels and semi-supervised learning on 2.5 million radiographs. While effective, their reliance on report-derived labels contrasts with our fully self-supervised approach that learns directly from multi-view radiographs without requiring textual annotations.

\section{Approach}
\subsection{Background}
\label{lab:sec_backgroud}
DINO (\textbf{Di}stillation with \textbf{No} Labels) \cite{caron2021emerging} and DINOv2 \cite{oquab2024dinov2} build upon the principles of self-supervised learning by leveraging Vision Transformers (ViT) \cite{dosovitskiy2021} and an innovative approach to knowledge distillation. Central to DINO's framework is its strategy of learning from multiple, randomly generated crops applied to each image. This enables the model to develop representations that are robust to variations in viewpoint and local image content, achieving a form of transformation invariance.

In the DINO framework, the input image undergoes a process of extensive data augmentation, producing multiple views of the same image. This augmentation strategy, often referred to as ``multi-crop," generates two types of views: global views and local views. Global views are typically larger crops (e.g., 224x224 pixels) that encompass a significant portion of the original image. Local views are smaller crops (e.g., 98x98 pixels) that focus on specific regions. These crops are generated with varying resolutions and fields of view, and are created through random resizing, cropping, and other transformations such as random horizontal flipping, color jittering (adjusting brightness, contrast, saturation, and hue), and Gaussian blurring. This process simulates the variations that might be encountered in real-world images, such as changes in lighting, viewpoint, and partial occlusions. DINO utilizes a teacher-student training paradigm, a form of knowledge distillation. Both the teacher and student networks are Vision Transformers, but they are trained differently. The student network receives all views (both global and local) and is trained to predict the output of the teacher network. The teacher network, however, only receives the global views. The teacher's weights are an exponential moving average (EMA) of the student's weights, providing a more stable and higher-quality target for the student to learn from. This process forces the student to learn representations that are consistent across different views (both global and local) of the same image, effectively aligning its output with the teacher's more ``informed" representation derived from the global views. This enforced consistency of features across different, augmented views of the same image is a key factor in DINO's ability to learn meaningful, semantically rich representations without the need for labeled data.

Our proposed research, VET-DINO, builds upon this foundational concept, but instead of relying on artificially generated crops of a single image, it leverages the inherent multi-view nature of veterinary radiographic studies.

\subsection{Multi-view Distillation}
\begin{figure}[ht]
    \centering
    \includegraphics[width=\linewidth]{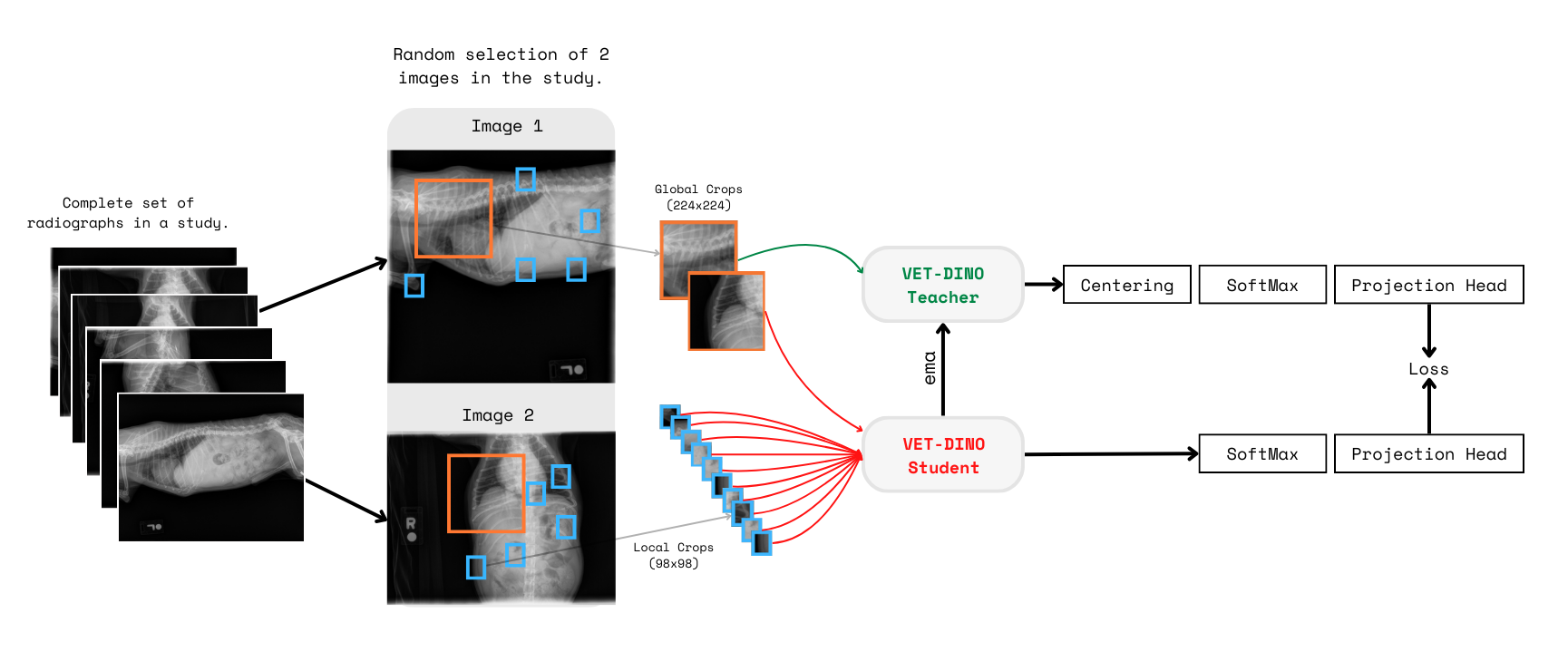} 
    \caption{Multi-view VET-DINO architecture. Two radiographic views are randomly selected from a single canine study (a set of radiographs from the same imaging session). From each view, two global crops and ten local crops are extracted and resized to 224x224 pixels and 98x98 pixels, respectively.  All crops are passed to the student Vision Transformer (ViT) network, while only the global crops from one randomly chosen view are passed to the teacher ViT network. The student network is trained to match the output of the teacher network, which receives a more ``global" perspective. The teacher's weights are an exponential moving average (EMA) of the student's weights. This process enables VET-DINO to learn view-invariant anatomical representations without manual annotations.} 
    \label{fig:mvd_vet_dino}
\end{figure}

VET-DINO is trained using crops from two random images within the same study, unlike DINO, which uses crops from a single image. The goal is to enable the model to learn from a combination of crops that represent different points of view within the same study. Figure \ref{fig:mvd_vet_dino} highlights how each image randomly selected from the study contributes to the model learning process.

\subsection{Dataset}
\label{lab:sec_dataset}
The dataset we used to pretrain our models consists of 5 million X-ray images drawn from approximately 668,000 unique canine studies. Most images were archived as lossy JPEGs with a quality setting of 89 (on a scale of 0-100, where 100 represents the least compression and highest quality) with a fixed width of 1024 pixels. This level of compression introduces some minor image artifacts, but was deemed acceptable for the scale of the dataset and the intended pre-training task. 

The remaining images that were not archived as lossy JPEGs came from clinical workflows as DICOM files, down-sampled to 512 pixel height and converted to PNG. The different image resolutions reflect changes in acquisition equipment and storage practices over the 14-year period. All images preserved their original aspect ratios and included metadata from Antech Imaging Services (AIS). The dataset spans real clinical cases from more than 3,500 hospitals and clinics, covering 14 years. Before modeling, duplicate images and images with low information content (e.g., images that were almost entirely blank or contained only a small portion of the anatomy) were removed using ImageMagick. A separate convolutional neural network (CNN) model, was trained to filter out images with significant artifacts and irrelevant views.

\subsection{Implementation Details}
\label{lab:sec_implementation}
The model leverages the ViT-S/14 architecture \cite{dosovitskiy2021} and is trained using the AdamW optimizer \cite{loshchilov2019} with a batch size of $192$ on a single NVIDIA H100 GPU. The learning rate is linearly warmed up over the first $80,000$ steps, held constant at $5 \times 10^{-2}$ for the next $320,000$ steps, and then decayed to $1 \times 10^{-4}$, following a cosine annealing schedule \cite{loshchilov2017sgdr} for the remainder of the $1,400,000$ training steps, for 177 epochs. The projection layer remains frozen during the initial warm-up phase of training to enhance stability. Furthermore, the size of the projection layer has been reduced to $16,384$ compared to $65,536$ in the original DINO implementation. This reduction was made to decrease memory consumption and was found to have minimal impact on downstream task performance (see Section \ref{lab:sec_finetune}). The weight decay is fixed at $0.04$, DropPath is set to $0.1$, and all training is performed in 32-bit precision.

For data augmentation, we apply random horizontal flipping with a probability of 0.1. We also employ a multi-crop strategy similar to DINO \cite{caron2021emerging}. This involves generating two global views and ten local views from the two selected images. Global views are created by randomly cropping regions that cover 25\% to 100\% of the image area, then resizing these crops to 224x224 pixels. Local views are created by randomly cropping regions that cover 5\% to 25\% of the image area, also resizing them to 98x98 pixels. The 10 local crops and 2 global crops are generated such that half originate from one randomly selected image in the study, and the other half originate from the second randomly selected image, allowing crops to overlap.

\section{Main Results}
\subsection{Evaluation Protocols}
\label{lab:sec_eval_protocols}
We evaluate our pre-trained VET-DINO model using two complementary approaches: k-Nearest Neighbors (k-NN) with a frozen feature extractor \cite{mathilde2020} and full fine-tuning \cite{Krizhevsky2012}. Both protocols are applied to a multi-label classification task of common radiographic findings using the evaluation dataset described in Section \ref{lab:sec_eval_dataset}.

The k-NN evaluation assesses representation quality in a non-parametric setting by classifying each image based on its k nearest neighbors in the embedding space generated by the frozen pre-trained encoder. For this protocol, we report F1-score, precision, and recall metrics. The optimal value of k was determined using GridSearch with 5-fold cross-validation, selecting the value that maximized the F1 score. 

For fine-tuning evaluation, we initialize the network with pre-trained student weights and optimize all parameters for the downstream task. Performance is measured using Area Under the ROC Curve (ROC AUC) and Average Precision (Avg Prec).

We compare VET-DINO against two baselines:
\begin{enumerate}
    \item Single-view VET-DINO: A DINOv2 model trained on the same dataset but using only one image per study (details in Appendix Section \ref{fig:svd_vet_dino})
    \item Standard DINOv2: The publicly available ImageNet pre-trained model from Meta
\end{enumerate}

\subsection{Evaluation Dataset}
\label{lab:sec_eval_dataset}
The evaluation dataset (Table \ref{tab:eval_df}) comprises a total of $177,526$ labeled canine radiographic images, selected from the larger pre-training dataset described in Section \ref{lab:sec_dataset}. Each image is annotated in a multi-label setting with findings spanning $47$ distinct radiographic abnormalities, encompassing a range of conditions affecting various anatomical systems. A complete list of these findings is provided in Appendix Section \ref{appdx:eval_count_all}. These labels were obtained through manual annotation by board-certified veterinary radiologists.

The dataset is divided into a training set of $100,285$ images and a validation set of $77,241$ images. This split was performed randomly, while ensuring a similar distribution of labels across both sets. To prevent data leakage, images from the same study were kept within the same split. Table \ref{tab:eval_df} provides the distribution of labels within each split. 

The relatively large proportion of data allocated to the validation set is deliberate and serves two key purposes. First, this allows for robust and statistically reliable evaluation of model performance, which is crucial for benchmarking against our other experiments, specifically the comparisons with the ImageNet pre-trained DINOv2 and the single-view baseline (as described in Section \ref{lab:sec_eval_protocols}). Second, the size of the training subset was chosen based on preliminary experiments demonstrating that this amount of training data was sufficient to achieve strong performance. By maintaining a consistent training set size across different experimental conditions, we can more confidently attribute performance differences to changes in model architecture or training strategies, rather than variations in the training data itself.

\begin{table}[]
\small
\centering

\resizebox{\textwidth}{!}{
    \begin{tabular}{l|r|r|r|r}
    \hline
    \multirow{2}{*}{\textbf{Label}}  & \multicolumn{2}{c|}{\textbf{Train Dataset}} & \multicolumn{2}{c}{\textbf{Validation Dataset}}  \\
    \cline{2-5}
    & \textbf{Negative} & \textbf{Positive}   & \textbf{Negative}  &  \textbf{Positive}  \\
    \hline
    Decreased serosal detail  & 96,922 (97\%)        & 3,363 (3\%)         & 74,631 (97\%)           & 2,610 (3\%)           \\
    Degenerative Joint Disease    & 91,775 (91\%)        & 8,510 (9\%)         & 70,478 (90\%)           & 6,763 (10\%)          \\
    Fat Opacity Mass (e.g. lipoma)  & 98,507 (98\%)        & 1,778 (2\%)         & 75,848 (98\%)           & 1,393 (2\%)           \\
    Mediastinal Mass Effect  & 99,599 (99\%)        & 686 (1\%)           & 76,667 (99\%)           & 574 (1\%)             \\
    Mediastinal Widening   & 99,617 (99\%)        & 668 (1\%)           & 76,695 (99\%)           & 546 (1\%)             \\
    Prostatic Enlargement & 99,736 (99\%)        & 549 (1\%)           & 76,863 (99\%)           & 378 (1\%)             \\
    Sign(s) of IVDD   & 94,000 (93\%)        & 6,285 (7\%)         & 72,520 (93\%)           & 4,721 (7\%)           \\
    Small Intestinal Obstruction  & 99,107 (99\%)        & 1,178 (1\%)         & 76,287 (99\%)           & 954 (1\%)             \\
    Small Kidney     & 98,057 (98\%)        & 2,228 (2\%)         & 75,527 (98\%)           & 1,714 (2\%)           \\
    Stifle Effusion   & 98,323 (98\%)        & 1,962 (2\%)         & 75,520 (98\%)           & 1,721 (2\%)           \\
    Subcutaneous Mass   & 98,846 (99\%)        & 1,439 (1\%)         & 76,097 (98\%)           & 1,144 (2\%)           \\
    Uterine Enlargement  & 99,968 (99\%)        & 317 (1\%)           & 76,965 (99\%)           & 276 (1\%)             \\
    \hline
    \end{tabular}
}
\caption{Training and validation dataset counts for a subset of 11 labels. To view counts for all 47 labels, please refer to Appendix Section \ref{appdx:eval_count_all}.}
\label{tab:eval_df}
\end{table}

\begin{figure}[h!]
    \centering
    \begin{minipage}[b]{0.2\textwidth}
        \includegraphics[width=\textwidth]{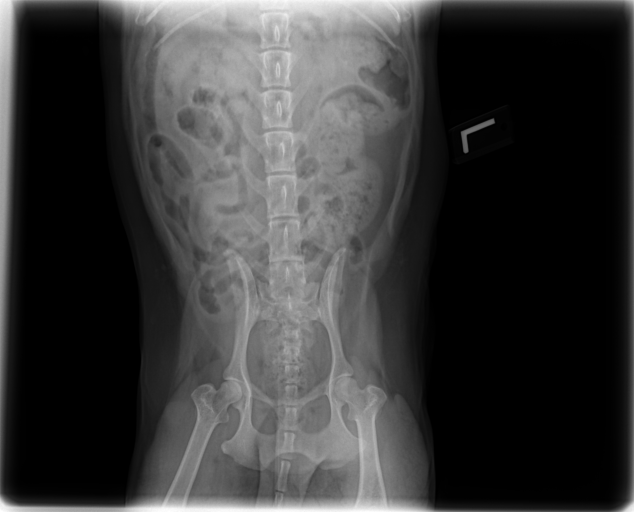}
    \end{minipage}
    \hspace{0.1\textwidth}
    \begin{minipage}[b]{0.2\textwidth}
        \includegraphics[width=\textwidth]{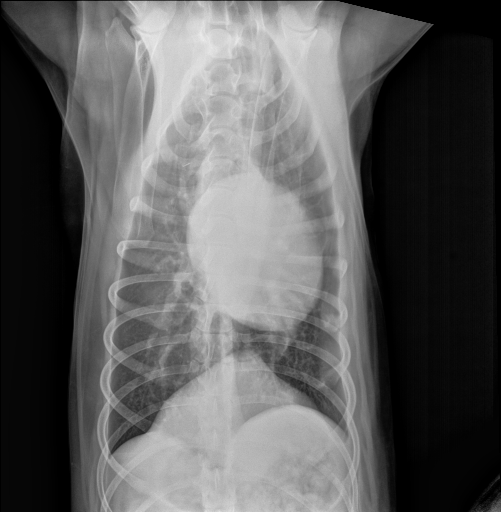}
    \end{minipage}
    
    \vspace{0.3cm} 
    \begin{minipage}[b]{0.2\textwidth}
        \centering
        \includegraphics[width=\textwidth]{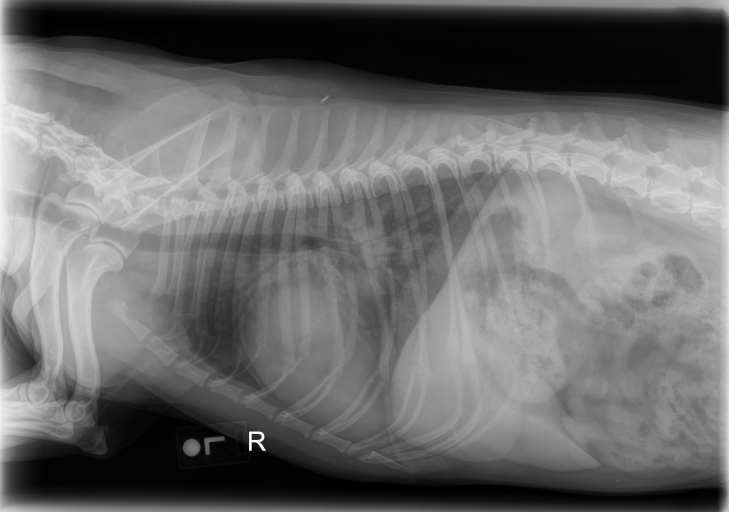}
    \end{minipage}
    \hspace{0.1\textwidth}
    \begin{minipage}[b]{0.2\textwidth}
        \centering
        \includegraphics[width=\textwidth]{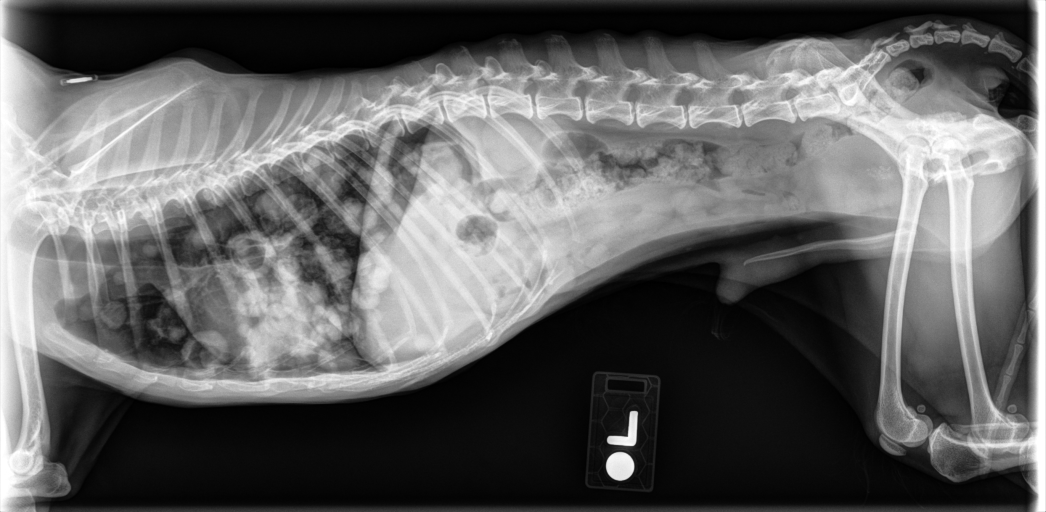}
    \end{minipage}

    \caption{Samples of radiographs from the validation dataset.} 
    \label{fig:sample_images}
\end{figure}
\FloatBarrier

\subsection{K-NN Classification Results}
\label{lab:sec_knn}
To evaluate the learned representations, we extracted features from both the training and validation/test sets using the frozen, pre-trained VET-DINO encoder.  Then a k-NN classifier (with k = 3, obtained using the approach described in Section \ref{lab:sec_eval_protocols}) was applied to the extracted features. The classifier was fit using the training set features and corresponding labels, and its performance was evaluated on the validation set features, using F1-score, precision, and recall as metrics. Table \ref{tab:knn_results} summarizes these results, comparing Multi-view VET-DINO with our Single-view VET-DINO baseline and the ImageNet pre-trained DINOv2 model. The k-nn results are reported on three gastrointestinal labels: Ingesta in the Stomach, indicating the presence of food material in the stomach; Gastric Foreign Materials (Debris), representing non-food substances within the stomach; and Foreign Body in the Small Intestines, identifying abnormal objects causing potential obstruction in the intestinal tract.

\begin{table}[ht]
    \centering
    \label{tab:combined_model_comparison}
    \resizebox{\textwidth}{!}{ 
    \begin{tabular}{l|l|c|c|c|c|c|c|c|c|c}
        \hline
        \multirow{3}{*}{\textbf{Architecture}} & \multirow{3}{*}{\textbf{Fine-tuning}} & \multicolumn{3}{c|}{\textbf{Ingesta in Stomach}} & \multicolumn{3}{c|}{\textbf{Gastric Foreign Material}} & \multicolumn{3}{c}{\textbf{Foreign Body in Small Intestine}} \\
        \cline{3-11}
        & & \textbf{F1} & \textbf{Precision} & \textbf{Recall} & \textbf{F1} & \textbf{Precision} & \textbf{Recall} & \textbf{F1} & \textbf{Precision} & \textbf{Recall} \\
        \hline
         VET-DINO       & Multi-image Studies            & \textbf{0.5733}    & \textbf{0.5089}    & \textbf{0.6564}    & \textbf{0.2583} & \textbf{0.2609}  & \textbf{0.3086} & \textbf{0.3517}  & \textbf{0.2748}   & \textbf{0.1272}\\
        VET-DINO (single view)        & Single-image Studies            & 0.2795             & 0.2630             & 0.2983       & 0.1413  & 0.1402   & 0.1425   & 0.1005 & 0.1098  & 0.0927    \\
        DINOv2         & None         & 0.2843             & 0.2717             & 0.2982     & 0.1338 & 0.1412  & 0.1272   & 0.0738 & 0.0833  &  0.06623    \\
        \hline
    \end{tabular}
    }
    \caption{Comparison of F1, Precision, and Recall scores from the k-NN classification for all models.}
    \label{tab:knn_results}
\end{table}

\subsection{Fine-tuning Results}
\label{lab:sec_finetune}
A ViT-Tiny model (vit-tiny-patch16-224, pretrained=False) from the timm library \cite{{wightman2019timm}} (based on the original Vision Transformer (ViT) architecture \cite{dosovitskiy2021}) is used as the classification layer and was added on top of the pre-trained, unfrozen VET-DINO encoder. The model was trained end-to-end using the AdamW optimizer \cite{loshchilov2019} with the learning rate set to $5 \times 10^{-6}$, a batch size of 512, and weight decay equal to 0.001. Training was conducted with 16-bit precision, employing a cosine annealing learning rate scheduler \cite{loshchilov2017sgdr}.  We used a binary cross-entropy (BCE) loss function, as this is a multi-label classification problem. The following data augmentations were applied during fine-tuning: resizing to 224x224 pixels, random horizontal flips with a probability of 0.5, random rotations within the range of ±30 degrees, random affine transformations with translation (10\% in both x and y directions), scaling (0.8 to 1.2), shear (±10 degrees), random perspective distortion with scale of 0.5 and probability of 0.5, and RandAugment \cite{cubuk2019randaug} with parameters N=3 (number of operations) and M=6 (magnitude of transformations). 

As a baseline, we also fine-tuned the publicly available, ImageNet pre-trained DINOv2 ViT-S/14 model from Meta \cite{oquab2024dinov2} on the same evaluation dataset.  To ensure a fair comparison, we used precisely the same hyperparameters, data augmentations, and training procedures for Multi-view VET-DINO, Single-view VET-DINO and the ImageNet pre-trained DINOv2 model.  The results, including Average Precision (Avg Prec) and ROC AUC, are summarized in Table \ref{tab:fiinetune_results}.


\begin{table}[h]
\centering
\resizebox{\textwidth}{!}{
    \begin{tabular}{l|r|r|r|r|r|r}
    \hline
    \multirow{2}{*}{\textbf{Label}}  & \multicolumn{2}{c|}{\textbf{Multi-view VET-DINO}} & \multicolumn{2}{c|}{\textbf{Single-view VET-DINO}} & \multicolumn{2}{c}{\textbf{DINOv2}} \\
    \cline{2-7}
     & \textbf{Avg Prec} & \textbf{ROC AUC} & \textbf{Avg Prec} & \textbf{ROC AUC}  & \textbf{Avg Prec} & \textbf{ROC AUC} \\
    \hline
    Decreased serosal detail   & \textbf{0.700}   & \textbf{0.961}  & 0.619  & 0.944  & 0.622  & 0.944  \\
    Degenerative Joint Disease  & \textbf{0.589} & \textbf{0.916}   & 0.466  & 0.877  & 0.484   & 0.881   \\
    Mediastinal Mass Effect  & \textbf{0.402} & \textbf{0.956}  & 0.371  & 0.948   & 0.393  & 0.952 \\
    Mediastinal Widening  & \textbf{0.521}  & \textbf{0.978}  & 0.484   & 0.968  & 0.521   & 0.971  \\
    Prostatic Enlargement & \textbf{0.358}  & \textbf{0.964}  & 0.277    & 0.948  & 0.326   & 0.944  \\
    Sign(s) of IVDD  & \textbf{0.622}   & \textbf{0.936}  & 0.541   & 0.912  & 0.563  & 0.919   \\
    Small Intestinal Obstruction  & \textbf{0.440}  & \textbf{0.960}  & 0.383   & 0.944   & 0.433  & 0.954 \\
    Small Kidney & \textbf{0.429}   & \textbf{0.963}   & 0.400   & 0.961  & 0.402   & 0.960  \\
    Stifle Effusion  & \textbf{0.785}  & \textbf{0.989}   & 0.753   & 0.989   & 0.771  & 0.989  \\
    Subcutaneous Mass   & \textbf{0.538} & \textbf{0.949}  & 0.476   & 0.931   & 0.509  & 0.889  \\
    Uterine Enlargement  & \textbf{0.590}  & \textbf{0.976}  & 0.415   & 0.933  & 0.535   & 0.949  \\
    \hline
    \end{tabular}
}
\caption{Fine-tuning Performance for 11 labels. To view results for all 47 labels, please refer to Appendix Section \ref{appdx:finetune_all}.}
\label{tab:fiinetune_results}
\end{table}

\subsection{Investigation of Anatomical Understanding}
\label{lab:sec_analysis}
Radiographs, by their nature, present a flattened, 2D representation of a complex 3D structure, requiring clinicians to mentally reconstruct the 3D anatomy. This is in contrast to modalities like computed tomography (CT), which directly generate 3D representations.  The ability to infer 3D information from 2D projections is particularly challenging in veterinary radiology due to significant variations in patient positioning, size, and breed-specific anatomical differences. This section investigates the extent to which VET-DINO achieves this goal of learning view-invariant representations that encode 3D information, facilitating generalization across different projections, and mitigating the inherent limitations of 2D radiographic data. To explore this, we analyze the attention maps from each model (Multi-view VET-DINO, Single-view VET-DINO, and untuned DINOv2) and visualize the embedding space of various anatomical structures.

\begin{figure}[h]
    \centering
    \includegraphics[width=\linewidth]{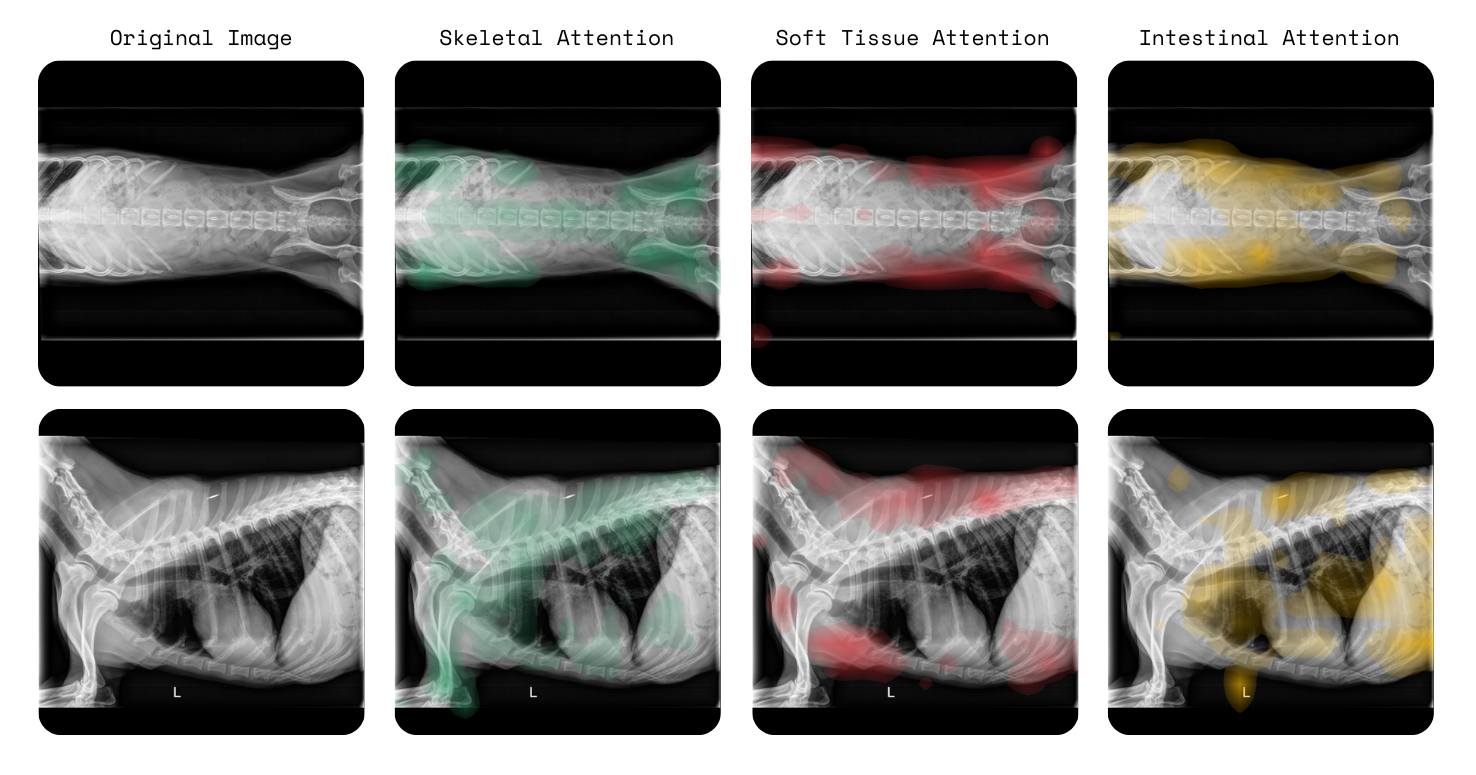} 
    \caption{The above figure displays ventrodorsal and left lateral radiographic projections of a canine patient acquired during the same visit. Superimposed on these images are attention maps derived from the final block head of the Multi-view VET-DINO model, illustrating the model's focus on specific anatomical regions. Notably, distinct attention heads appear to be independently attending to the skeletal system (green), the soft tissue/muscular system (red), and the gastrointestinal system (yellow). This observation suggests the model exhibits view-invariant attention to anatomical structures although imperfectly, highlighting its capacity to learn consistent representations of anatomical features across different radiographic views.}
    \label{fig:mult_attn}
\end{figure}

\subsubsection{Attention Visualization}
\label{lab:sec_attn_viz}
\begin{figure}[h]
    \centering
    \includegraphics[width=\linewidth]{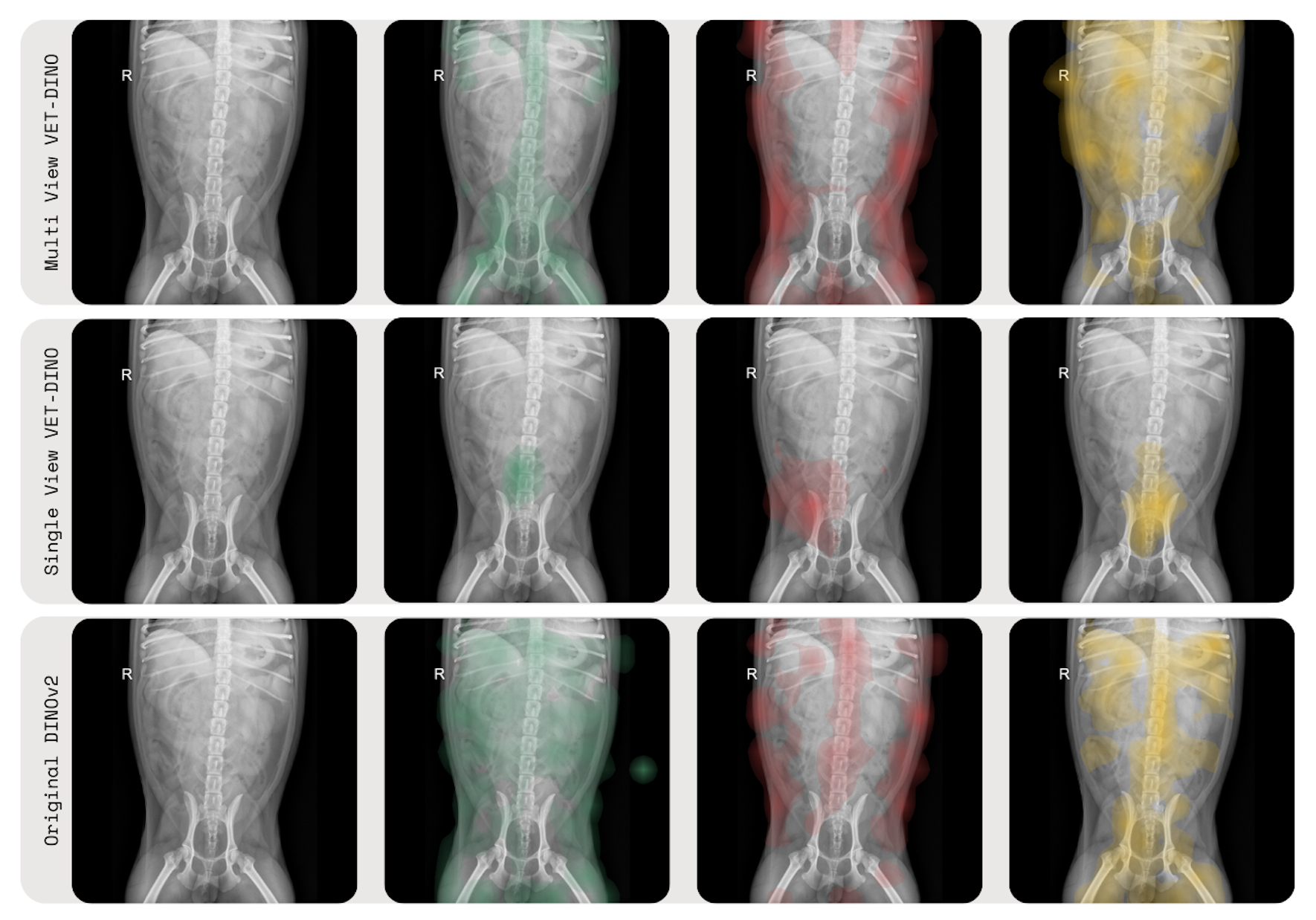} 
    \caption{Comparative analysis of the Multi-view VET-DINO model against a Single-view VET-DINO model and the original DINOv2, using identical radiographic input, demonstrates the superior ability of the Multi-view VET-DINO model to attend to relevant anatomical structures in canine patients. We see that the Multi-view VET-DINO effectively attends to the skeletal system (green), soft tissue (red), and intestinal system (yellow). This observation underscores the  hypothesis that fine-tuning with multi-view radiographic studies enhances the model's learning capacity and facilitates the development of more robust and comprehensive anatomical representations.}
    \label{fig:mult_attn2}
\end{figure}

Figure \ref{fig:mult_attn} displays representative attention maps from the student network's final transformer block's attention head of the VET-DINO model. These maps were generated from radiographs of the same canine study, including lateral and ventrodorsal projections. The VET-DINO attention maps consistently highlight regions corresponding to the skeletal system, particularly the spine and pelvis, across different views. This focused attention suggests that the model has learned to identify these key anatomical structures as consistent features, regardless of their 2D projection. In contrast, as shown in Figure \ref{fig:mult_attn2}, the attention maps from the Single-view VET-DINO and the ImageNet pre-trained DINOv2 models exhibit a more dispersed pattern, with less consistent focus on specific anatomical structures. This difference in attention patterns suggests that VET-DINO's multi-view training leads to a greater emphasis on anatomically relevant features.

\FloatBarrier

\subsubsection{Patch Embedding Similarity}
\label{lab:sec_pat_em}
\begin{figure}[h]
    \centering
    \includegraphics[width=\linewidth]{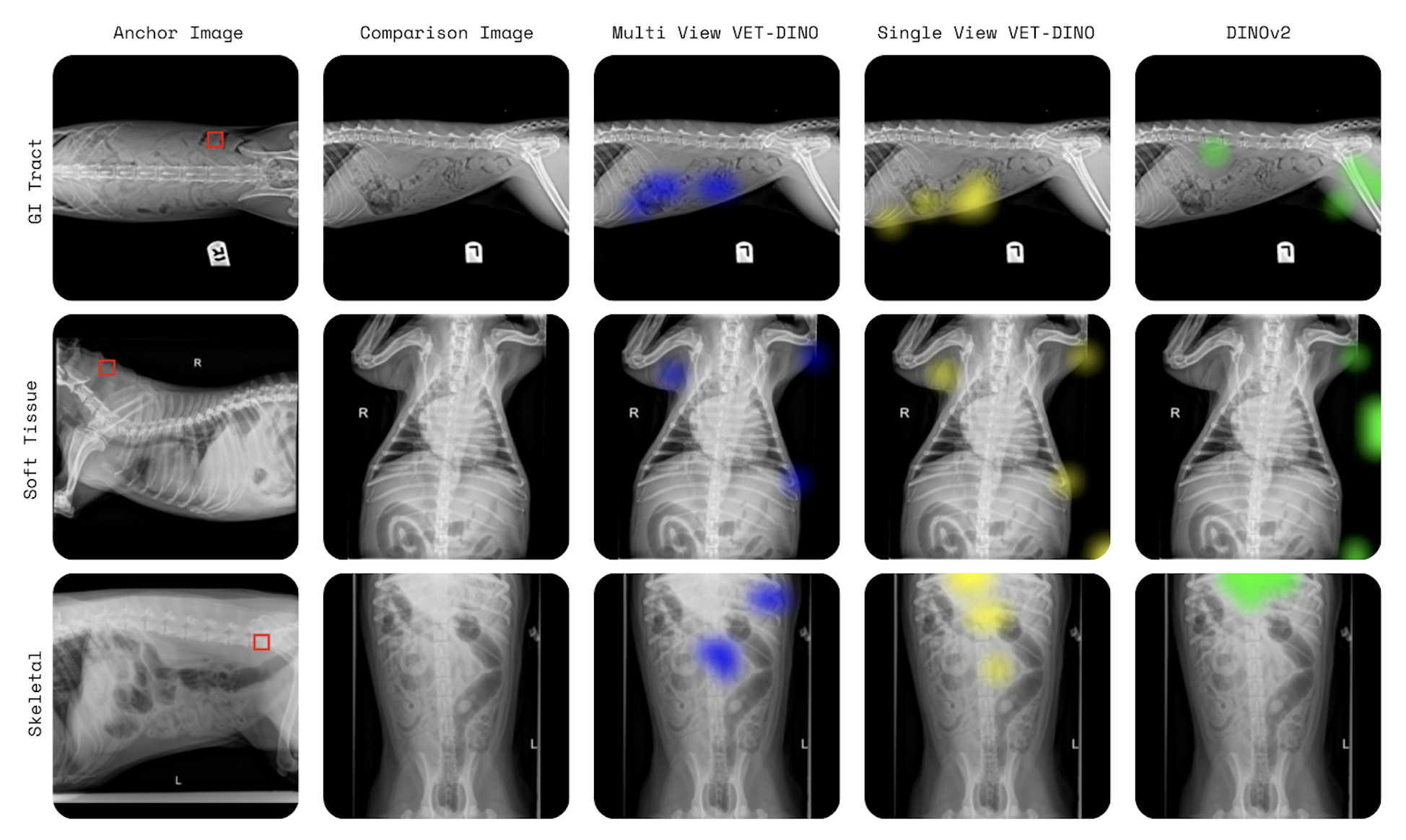} 
    \caption{Visualization of cosine similarity between patch embeddings for a multi-view VET-DINO model, single-view VET-DINO, and untuned DINOv2. The anchor patch (red box) and comparison image originate from the same patient and radiographic session. Top-5 most similar patches are highlighted, demonstrating superior view-invariant anatomical feature identification in the multi-view model, as compared to the single-view model and original DINOv2.}
    \label{fig:cosine_sim}
\end{figure}

To quantify the observed consistency of attention on key anatomical structures, we analyzed the cosine similarity between patch embeddings from different radiographic views of the same study (Figure \ref{fig:cosine_sim}). Specifically, we manually selected patches corresponding to three anatomical regions: the gastrointestinal tract, soft tissue/muscular structures, and the skeletal system. For each anchor patch in a selected image, we calculated the cosine similarity between its embedding and the embeddings of all patches in the accompanying image within the same study. High cosine similarity indicates that the model represents the same anatomical region similarly, despite differences in its 2D projection. Across a small sample of 24 image pairs, we calculated an average cosine similarity of 0.98 between the anchor patch and the top 5 most similar patches in the comparison image. This suggests that VET-DINO learns representations that exhibit a degree of view-invariance.

Patch embeddings are extracted from the output of the VET-DINO student ViT encoder's final transformer block, before the [CLS] token embedding. Specifically, for each image, the activations from this layer are reshaped into a set of patch embeddings, each corresponding to a 14x14 pixel region in the input image (given the ViT-S/14 architecture). To quantify inter-view similarity, an anchor patch is selected from one view of a study. The cosine similarity is then calculated between the anchor patch embedding and all patch embeddings from a different, comparison view of the same study. This provides a measure of the similarity between local image features across different radiographic projections. A visual illustration of this process is provided in Appendix \ref{appdx:cos_sim}, Figure \ref{fig:cos_sim_steps}.

\section{Conclusion, Limitations, and Future Work}

In this paper, we introduced VET-DINO, a novel self-supervised learning framework that leverages the inherent multi-view nature of veterinary radiographic studies to learn view-invariant anatomical representations. Unlike traditional approaches that rely on single-image augmentations, VET-DINO uses actual radiographic views from the same study as natural augmentations, enabling the model to learn representations that are less sensitive to variations in patient orientation and projection angle. Our experiments on a large-scale dataset of 5 million canine radiographs demonstrate that VET-DINO trained on multi-view studies outperforms both its single-view complement and an ImageNet pre-trained DINOv2 model on downstream veterinary imaging tasks, including k-NN classification and fine-tuning for abnormality detection. Furthermore, our analysis of attention maps and patch embedding similarities provides strong evidence that VET-DINO learns to focus on key anatomical structures consistently across different views, suggesting the development of representations that capture a degree of 3D anatomical information from 2D projections.

Despite these promising results, this work has limitations. First, our evaluation focused primarily on 47 findings in canine radiographs. Further evaluation on a broader range of anatomical regions, species, and clinical conditions is necessary to fully assess the generalization of VET-DINO. Second, while our cosine similarity analysis suggests view-invariance, it relies on manually selected anatomical regions. Future work should explore more automated and objective methods for assessing this property. Finally, the current study utilizes a relatively small ViT-S/14 architecture.

Future research will focus on several key areas:
\begin{itemize}
\item {\bf Scaling}: We will explore larger model architectures (e.g., ViT-L) and significantly expanded datasets, including a wider range of radiographic studies and both canine and feline patients. The aim will be to improve model performance and generalization.
\item {\bf 3D Understanding}: We will investigate methods for explicitly reconstructing or inferring 3D anatomical information from the 2D representations learned by VET-DINO.  This could involve training a separate module to predict 3D landmark locations from the 2D embeddings and potentially exploring techniques from neural rendering to generate novel views.
\item {\bf Domain Knowledge Integration}:  We will incorporate domain-specific knowledge from veterinary radiology to further enhance the model's performance and interpretability. This could include incorporating anatomical priors into the loss function and using more expert-labeled data for semi-supervised learning.
\end{itemize}

Ultimately, this research aims to develop robust, interpretable, and clinically useful self-supervised learning models for veterinary medical imaging. By leveraging the inherent multi-view nature of radiographic data and incorporating domain expertise, we believe this work has the potential to significantly improve diagnostic accuracy, aid in clinical decision-making, and ultimately enhance animal healthcare.

\section{Acknowledgments}
The authors express their sincere appreciation to Conrad Stack, Emil Walleser, and Jeff Breeding-Allison for their valuable insights and expertise throughout this research. Their contributions to evaluating VET-DINO outputs and expertise in veterinary radiology were essential to the success of this project. A special thanks is extended to Noel Codella for his advice on \ref{lab:sec_pat_em}.

\newpage
\bibliographystyle{plain}
\bibliography{sample}

\begin{thebibliography}{10}

\bibitem{bannur2024maira}
Shruthi Bannur, Kenza Bouzid, Daniel~C Castro, Anton Schwaighofer, Anja Thieme, Sam Bond-Taylor, Maximilian Ilse, Fernando P{\'e}rez-Garc{\'\i}a, Valentina Salvatelli, Harshita Sharma, et~al.
\newblock Maira-2: Grounded radiology report generation.
\newblock {\em arXiv preprint arXiv:2406.04449}, 2024.

\bibitem{bertrand2019lateral}
Hadrien Bertrand, Mohammad Hashir, and Joseph~Paul Cohen.
\newblock Do lateral views help automated chest x-ray predictions?
\newblock {\em arXiv preprint arXiv:1904.08534}, 2019.

\bibitem{burti2024artificial}
Silvia Burti, Tommaso Banzato, Simon Coghlan, Marek Wodziniski, Margherita Bendazzoli, and Alessandro Zotti.
\newblock Artificial intelligence in veterinary diagnostic imaging: Perspectives and limitations.
\newblock {\em Research in Veterinary Science}, page 105317, 2024.

\bibitem{mathilde2020}
Mathilde Caron, Ishan Misra, Julien Mairal, Priya Goyal, Piotr Bojanowski, and Armand Joulin.
\newblock Unsupervised learning of visual features by contrasting cluster assignments.
\newblock {\em CoRR}, abs/2006.09882, 2020.

\bibitem{caron2021emerging}
Mathilde Caron, Hugo Touvron, Ishan Misra, Herv{\'e} J{\'e}gou, Julien Mairal, Piotr Bojanowski, and Armand Joulin.
\newblock Emerging properties in self-supervised vision transformers.
\newblock In {\em Proceedings of the IEEE/CVF international conference on computer vision}, pages 9650--9660, 2021.

\bibitem{chen2019med3d}
Sihong Chen, Kai Ma, and Yefeng Zheng.
\newblock Med3d: Transfer learning for 3d medical image analysis.
\newblock {\em arXiv preprint arXiv:1904.00625}, 2019.

\bibitem{chen2020simple}
Ting Chen, Simon Kornblith, Mohammad Norouzi, and Geoffrey Hinton.
\newblock A simple framework for contrastive learning of visual representations.
\newblock In {\em International conference on machine learning}, pages 1597--1607. PMLR, 2020.

\bibitem{cubuk2019randaug}
Ekin~D. Cubuk, Barret Zoph, Jonathon Shlens, and Quoc~V. Le.
\newblock Randaugment: Practical automated data augmentation with a reduced search space, 2019.

\bibitem{das2023viewclr}
Srijan Das and Michael~S Ryoo.
\newblock Viewclr: Learning self-supervised video representation for unseen viewpoints.
\newblock In {\em Proceedings of the IEEE/CVF Winter Conference on Applications of Computer Vision}, pages 5573--5583, 2023.

\bibitem{dosovitskiy2021}
Alexey Dosovitskiy, Lucas Beyer, Alexander Kolesnikov, Dirk Weissenborn, Xiaohua Zhai, Thomas Unterthiner, Mostafa Dehghani, Matthias Minderer, Georg Heigold, Sylvain Gelly, Jakob Uszkoreit, and Neil Houlsby.
\newblock An image is worth 16x16 words: Transformers for image recognition at scale, 2021.

\bibitem{fitzke2021rapidread}
Michael Fitzke, Conrad Stack, Andre Dourson, Rodrigo M.~B. Santana, Diane Wilson, Lisa Ziemer, Arjun Soin, Matthew~P. Lungren, Paul Fisher, and Mark Parkinson.
\newblock Rapidread: Global deployment of state-of-the-art radiology ai for a large veterinary teleradiology practice, 2021.

\bibitem{he2020momentum}
Kaiming He, Haoqi Fan, Yuxin Wu, Saining Xie, and Ross Girshick.
\newblock Momentum contrast for unsupervised visual representation learning.
\newblock In {\em Proceedings of the IEEE/CVF conference on computer vision and pattern recognition}, pages 9729--9738, 2020.

\bibitem{hennessey2022artificial}
Erin Hennessey, Matthew DiFazio, Ryan Hennessey, and Nicky Cassel.
\newblock Artificial intelligence in veterinary diagnostic imaging: A literature review.
\newblock {\em Veterinary Radiology \& Ultrasound}, 63:851--870, 2022.

\bibitem{Krizhevsky2012}
Alex Krizhevsky, Ilya Sutskever, and Geoffrey~E Hinton.
\newblock Imagenet classification with deep convolutional neural networks.
\newblock In F.~Pereira, C.J. Burges, L.~Bottou, and K.Q. Weinberger, editors, {\em Advances in Neural Information Processing Systems}, volume~25. Curran Associates, Inc., 2012.

\bibitem{loshchilov2017sgdr}
Ilya Loshchilov and Frank Hutter.
\newblock Sgdr: Stochastic gradient descent with warm restarts, 2017.

\bibitem{loshchilov2019}
Ilya Loshchilov and Frank Hutter.
\newblock Decoupled weight decay regularization, 2019.

\bibitem{ober2006comparison}
Christopher~P Ober and Don Barber.
\newblock Comparison of two-vs. three-view thoracic radiographic studies on conspicuity of structured interstitial patterns in dogs.
\newblock {\em Veterinary Radiology \& Ultrasound}, 47(6):542--545, 2006.

\bibitem{oquab2024dinov2}
Maxime Oquab, Timothée Darcet, Théo Moutakanni, Huy Vo, Marc Szafraniec, Vasil Khalidov, Pierre Fernandez, Daniel Haziza, Francisco Massa, Alaaeldin El-Nouby, Mahmoud Assran, Nicolas Ballas, Wojciech Galuba, Russell Howes, Po-Yao Huang, Shang-Wen Li, Ishan Misra, Michael Rabbat, Vasu Sharma, Gabriel Synnaeve, Hu~Xu, Hervé Jegou, Julien Mairal, Patrick Labatut, Armand Joulin, and Piotr Bojanowski.
\newblock Dinov2: Learning robust visual features without supervision, 2024.

\bibitem{perez2024rad}
Fernando P{\'e}rez-Garc{\'\i}a, Harshita Sharma, Sam Bond-Taylor, Kenza Bouzid, Valentina Salvatelli, Maximilian Ilse, Shruthi Bannur, Daniel~C Castro, Anton Schwaighofer, Matthew~P Lungren, et~al.
\newblock Rad-dino: Exploring scalable medical image encoders beyond text supervision.
\newblock {\em arXiv preprint arXiv:2401.10815}, 2024.

\bibitem{rajpurkar2017chexnet}
P~Rajpurkar.
\newblock Chexnet: Radiologist-level pneumonia detection on chest x-rays with deep learning.
\newblock {\em ArXiv abs/1711}, 5225, 2017.

\bibitem{schroff2015facenet}
Florian Schroff, Dmitry Kalenichenko, and James Philbin.
\newblock Facenet: A unified embedding for face recognition and clustering.
\newblock In {\em Proceedings of the IEEE conference on computer vision and pattern recognition}, pages 815--823, 2015.

\bibitem{shah2024mv2mae}
Ketul Shah, Robert Crandall, Jie Xu, Peng Zhou, Marian George, Mayank Bansal, and Rama Chellappa.
\newblock Mv2mae: Multi-view video masked autoencoders.
\newblock {\em arXiv preprint arXiv:2401.15900}, 2024.

\bibitem{vu2021medaug}
Yen Nhi~Truong Vu, Richard Wang, Niranjan Balachandar, Can Liu, Andrew~Y Ng, and Pranav Rajpurkar.
\newblock Medaug: Contrastive learning leveraging patient metadata improves representations for chest x-ray interpretation.
\newblock In {\em Machine Learning for Healthcare Conference}, pages 755--769. PMLR, 2021.

\bibitem{wannenmacher2023studyformer}
Lucas Wannenmacher, Michael Fitzke, Diane Wilson, and Andre Dourson.
\newblock Studyformer: Attention-based and dynamic multi view classifier for x-ray images.
\newblock {\em arXiv preprint arXiv:2302.11840}, 2023.

\bibitem{wightman2019timm}
Ross Wightman.
\newblock Pytorch image models.
\newblock \url{https://github.com/rwightman/pytorch-image-models}, 2019.

\bibitem{wilson2022role}
Diane~U Wilson, Michael~Q Bailey, and John Craig.
\newblock The role of artificial intelligence in clinical imaging and workflows.
\newblock {\em Veterinary Radiology \& Ultrasound}, 63:897--902, 2022.

\bibitem{zhou2021preservational}
Hong-Yu Zhou, Chixiang Lu, Sibei Yang, Xiaoguang Han, and Yizhou Yu.
\newblock Preservational learning improves self-supervised medical image models by reconstructing diverse contexts.
\newblock In {\em Proceedings of the IEEE/CVF International Conference on Computer Vision}, pages 3499--3509, 2021.

\end{thebibliography}

\newpage

\appendix
\section{Training Loss}
\begin{figure}[h]
    \centering
    \includegraphics[scale=0.5]{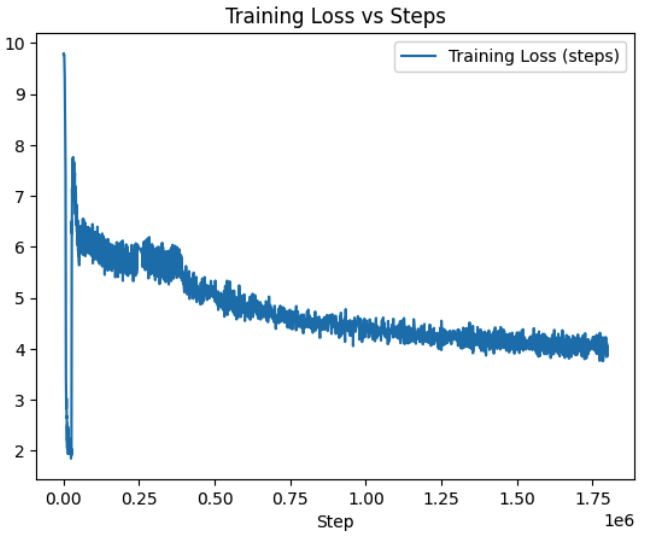}
    \caption{VET-DINO pre-training loss illustrates the training loss trajectory during the model’s self-training process. Across all our experiments, we consistently observed an abrupt collapse in the loss, followed by a recovery to a more conventional pattern.}
    \label{fig:training_loss}
\end{figure}
\FloatBarrier
\newpage
\section{Single-view VET-DINO}
\begin{figure}[h]
    \centering
    \includegraphics[width=\linewidth]{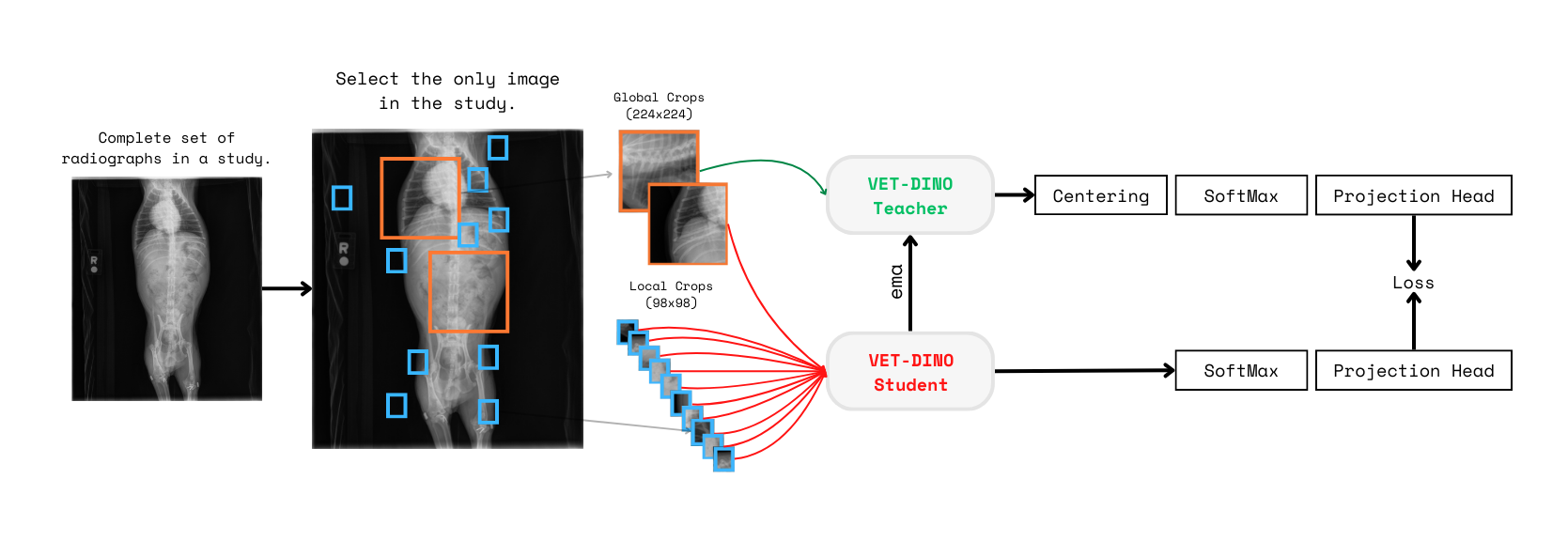}
    \caption{Single-view VET-DINO Architecture. Unlike Multi-view VET-DINO, a single radiographic view is taken from a single canine study. Two global crops and ten local crops are extracted and resized to 224x224 pixels and 98x98 pixels, respectively.  All crops are passed to the student Vision Transformer (ViT) network, while only one global crop is passed to the teacher ViT network. The student network is trained to match the output of the teacher network, which receives a more ``global" perspective. The teacher's weights are an exponential moving average (EMA) of the student's weights. This process enables VET-DINO to learn view-invariant anatomical representations without manual annotations.}
    \label{fig:svd_vet_dino}
\end{figure}
\FloatBarrier

\section{Cosine Similarity Computations}
\label{appdx:cos_sim}
\begin{figure}[h]
    \centering
    \includegraphics[width=\linewidth]{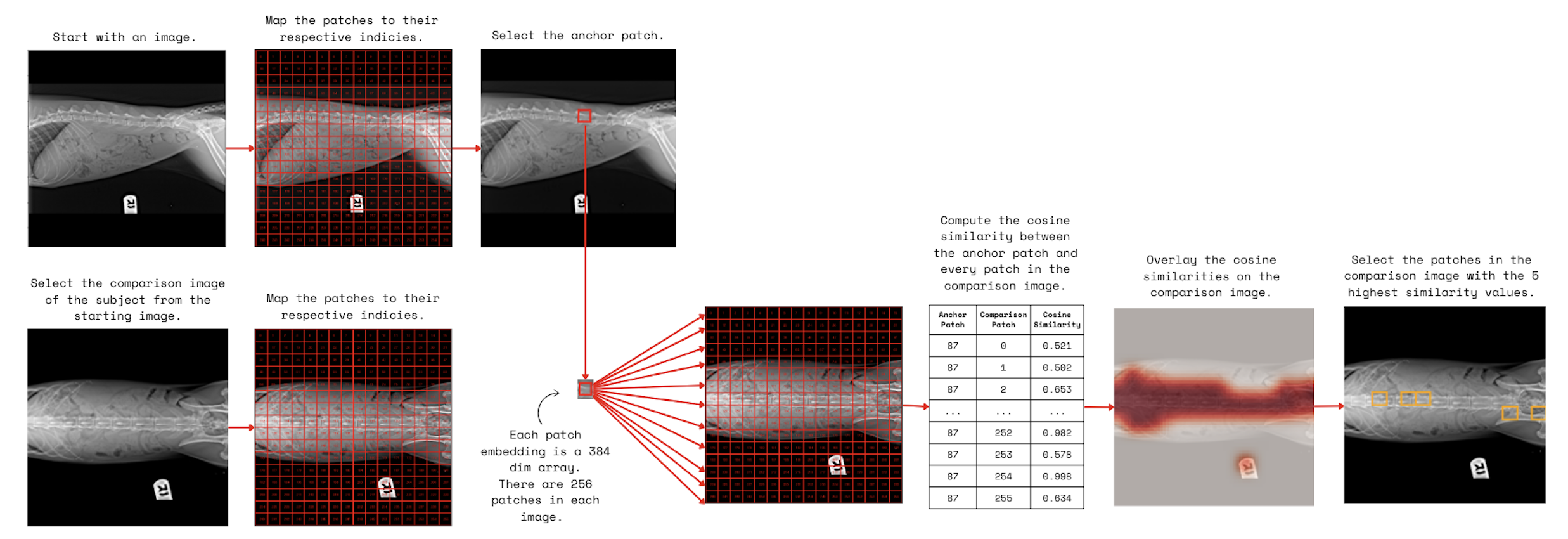} 
    \caption{Patch-level similarity is assessed by comparing embeddings extracted from the final layer activations of the ViT. The anchor patch embedding is contrasted with the embeddings of all patches in the comparison image. High cosine similarity scores between embeddings signify similar semantic content, as seen in the final image with the five most similar patches highlighted.}
    \label{fig:cos_sim_steps}
\end{figure}
\FloatBarrier

\section{Dataset Details for All Labels}
\begin{table}[]
\label{appdx:eval_count_all}
\small
\centering
\resizebox{\textwidth}{!}{
\begin{tabular}{l|r|r|r|r}
\hline
\multirow{2}{*}{\textbf{Label}}  & \multicolumn{2}{c|}{\textbf{Train Dataset}} & \multicolumn{2}{c}{\textbf{Validation Dataset}}  \\
\cline{2-5}
& \textbf{Negative} & \textbf{Positive}   & \textbf{Negative}  &  \textbf{Positive}  \\
\hline
Aggressive Bone Lesion                                   & 99,361 (99\%)        & 924 (1\%)           & 76,482 (99\%)           & 759 (1\%)             \\
Caudal Abdominal Mass                                    & 100,173 (99\%)       & 112 (1\%)           & 77,139 (99\%)           & 102 (1\%)             \\
Constipation Obstipation                                 & 99,783 (99\%)        & 502 (1\%)           & 76,822 (99\%)           & 419 (1\%)             \\
Cranial Abdominal Mass                                   & 99,885 (99\%)        & 400 (1\%)           & 76,957 (99\%)           & 284 (1\%)             \\
Decreased serosal detail                                 & 96,922 (97\%)        & 3,363 (3\%)         & 74,631 (97\%)           & 2,610 (3\%)           \\
Degenerative Joint Disease                               & 91,775 (91\%)        & 8,510 (9\%)         & 70,478 (90\%)           & 6,763 (10\%)          \\
Esophagal Dilation                                       & 98,916 (99\%)        & 1,369 (1\%)         & 76,214 (99\%)           & 1,027 (1\%)           \\
Fat Opacity Mass (e.g. lipoma)                           & 98,507 (98\%)        & 1,778 (2\%)         & 75,848 (98\%)           & 1,393 (2\%)           \\
Foreign Body in the Small Intestines                     & 99,329 (99\%)        & 956 (1\%)           & 76,508 (99\%)           & 733 (1\%)             \\
Gall Bladder Calculi                                     & 99,700 (99\%)        & 585 (1\%)           & 76,759 (99\%)           & 482 (1\%)             \\
Gastric Dilatation Volvulus                              & 100,195 (99\%)       & 90 (1\%)            & 77,164 (99\%)           & 77 (1\%)              \\
Gastric Distention                                       & 95,022 (94\%)        & 5,263 (6\%)         & 73,727 (95\%)           & 3,514 (5\%)           \\
Gastric Foreign Material (debris)                        & 95,489 (95\%)        & 4,796 (5\%)         & 74,241 (96\%)           & 3,000 (4\%)           \\
Hepatic Mineralization                                   & 99,883 (99\%)        & 402 (1\%)           & 76,948 (99\%)           & 293 (1\%)             \\
Ingesta in the stomach                                   & 86,000 (83\%)        & 14,285 (17\%)       & 73,646 (95\%)           & 3,595 (5\%)           \\
Irregular material in the small intestines               & 98,633 (92\%)        & 1,652 (2\%)         & 76,466 (99\%)           & 775 (1\%)             \\
Irregular small intestinal gas patterns                  & 99,991 (99\%)        & 294 (1\%)           & 77,095 (99\%)           & 146 (1\%)             \\
Large Kidney                                             & 100,088 (99\%)       & 197 (1\%)           & 77,073 (99\%)           & 168 (1\%)             \\
Limb Fracture                                            & 99,337 (99\%)        & 948 (1\%)           & 76,481 (99\%)           & 760 (1\%)             \\
Luxation                                                 & 99,068 (99\%)        & 1,217 (1\%)         & 76,263 (99\%)           & 978 (1\%)             \\
Mediastinal Mass Effect                                  & 99,599 (99\%)        & 686 (1\%)           & 76,667 (99\%)           & 574 (1\%)             \\
Mediastinal Widening                                     & 99,617 (99\%)        & 668 (1\%)           & 76,695 (99\%)           & 546 (1\%)             \\
Megacolon                                                & 100,095 (99\%)       & 190 (1\%)           & 77,126 (99\%)           & 115 (1\%)             \\
Mid Abdominal Mass                                       & 99,765 (99\%)        & 520 (1\%)           & 76,872 (99\%)           & 369(1\%)              \\
Misshapen Kidney(s)                                      & 99,862 (99\%)        & 423 (1\%)           & 76,945 (99\%)           & 296 (1\%)             \\
Pneumothorax                                             & 99,968 (99\%)        & 317 (1\%)           & 77,027 (99\%)           & 214 (1\%)             \\
Prostatic Enlargement                                    & 99,736 (99\%)        & 549 (1\%)           & 76,863 (99\%)           & 378 (1\%)             \\
Pulmonary Alveolar                                       & 94,904 (94\%)        & 5,381 (6\%)         & 72,919 (94\%)           & 4,322 (6\%)           \\
Pulmonary Interstitial Nodule(s)                         & 98,758 (98\%)        & 1,527 (2\%)         & 76,054 (98\%)           & 1,187 (2\%)           \\
Pulmonary Mass (Over 1 cm)                               & 98,665 (98\%)        & 1,620 (2\%)         & 75,970 (98\%)           & 1,271 (2\%)           \\
Pulmonary Vascular                                       & 99,021 (99\%)        & 1,264 (1\%)         & 76,232 (99\%)           & 1,009 (1\%)           \\
Pyloric outflow obstruction                              & 100,221 (99\%)       & 64(1\%)             & 77,207 (99\%)           & 34 (1\%)              \\
Renal Mineralization                                     & 98,989 (99\%)        & 1,296 (1\%)         & 76,288 (99\%)           & 953 (1\%)             \\
Rib Fracture                                             & 100,087 (99\%)       & 198 (1\%)           & 77,102 (99\%)           & 139 (1\%)             \\
Sign(s) of IVDD                                          & 94,000 (93\%)        & 6,285 (7\%)         & 72,520 (93\%)           & 4,721 (7\%)           \\
Sign(s) of Pleural Effusion                              & 98,681 (98\%)        & 1,604 (2\%)         & 75,918 (98\%)           & 1,323 (2\%)           \\
Small Intestinal Obstruction                             & 99,107 (99\%)        & 1,178 (1\%)         & 76,287 (99\%)           & 954 (1\%)             \\
Small Intestinal Plication                               & 100,116 (99\%)       & 169 (1\%)           & 77,121 (99\%)           & 120 (1\%)             \\
Small Kidney                                             & 98,057 (98\%)        & 2,228 (2\%)         & 75,527 (98\%)           & 1,714 (2\%)           \\
Small Liver                                              & 100,080 (99\%)       & 205 (1\%)           & 77,111 (99\%)           & 130 (1\%)             \\
Splenomegaly                                             & 99,081 (99\%)        & 1,204 (1\%)         & 76,322 (99\%)           & 919 (1\%)             \\
Sternal Lymph Node Enlargement                           & 99,896 (99\%)        & 389 (1\%)           & 76,902 (99\%)           & 339 (1\%)             \\
Stifle Effusion                                          & 98,323 (98\%)        & 1,962 (2\%)         & 75,520 (98\%)           & 1,721 (2\%)           \\
Subcutaneous Mass                                        & 98,846 (99\%)        & 1,439 (1\%)         & 76,097 (98\%)           & 1,144 (2\%)           \\
Subcutaneous Nodule                                      & 99,952 (99\%)        & 333 (1\%)           & 76,917 (99\%)           & 324 (1\%)             \\
Urinary Bladder Calculus Calculi                         & 99,282 (99\%)        & 1,003 (1\%)         & 76,478 (99\%)           & 763 (1\%)             \\
Uterine Enlargement                                      & 99,968 (99\%)        & 317 (1\%)           & 76,965 (99\%)           & 276 (1\%)             \\
\hline
\end{tabular}
}
\end{table}
\FloatBarrier

\section{Fine-tuning Results for All Labels}
\begin{table}[h]
\label{appdx:finetune_all}
\centering
\caption{Performance Metrics for All 47 Findings}
\resizebox{\textwidth}{!}{
    \begin{tabular}{l|r|r|r|r|r|r}
    \hline
    \multirow{2}{*}{\textbf{Label}}  & \multicolumn{2}{c|}{\textbf{Multi-view VET-DINO}} & \multicolumn{2}{c|}{\textbf{Single-view VET-DINO}} & \multicolumn{2}{c}{\textbf{DINOv2}} \\
    \cline{2-7}
     & \textbf{AP}        & \textbf{AUC}       & \textbf{AP}        & \textbf{AUC}        & \textbf{AP} & \textbf{AUC} \\
    \hline
    Aggressive   Bone Lesion                                 & \textbf{0.289} & \textbf{0.910} & 0.101 & 0.883 & 0.107 & 0.884 \\
    Caudal Abdominal Mass                                    & \textbf{0.029} & \textbf{0.871} & 0.016 & 0.866 & 0.020 & 0.839 \\
    Constipation/Obstipation                                 & \textbf{0.239} & \textbf{0.920} & 0.219 & 0.906 & 0.234 & 0.917 \\
    Cranial Abdominal Mass                                   & 0.158 & \textbf{0.894} & 0.124 & 0.873 & \textbf{0.172} & 0.864 \\
    Decreased serosal detail                                 & \textbf{0.700} & \textbf{0.961} & 0.619 & 0.944 & 0.622 & 0.944 \\
    Degenerative Joint   Disease                             & \textbf{0.589} & \textbf{0.916} & 0.466 & 0.877 & 0.484 & 0.881 \\
    Esophagal Dilation                                       & 0.380 & \textbf{0.921} & 0.298 & 0.898 & \textbf{0.391} & 0.920 \\
    Fat Opacity Mass (e.g.   lipoma)                         & \textbf{0.684} & \textbf{0.976} & 0.573 & 0.959 & 0.621 & 0.920 \\
    Foreign Body in the Small   Intestines                   & \textbf{0.220} & \textbf{0.930} & 0.192 & 0.914 & 0.216 & 0.927 \\
    Gall Bladder Calculi                                     & \textbf{0.172} & \textbf{0.904} & 0.115 & 0.887 & 0.172 & 0.877 \\
    Gastric Dilatation   Volvulus                            & 0.511 & \textbf{0.967} & \textbf{0.541} & 0.964 & 0.527 & 0.895 \\
    Gastric Distention                                       & 0.778 & \textbf{0.976} & 0.766 & 0.973 & \textbf{0.781} & 0.975 \\
    Gastric Foreign Material   (debris)                      & \textbf{0.502} & 0.901 & 0.463 & 0.885 & 0.482 & \textbf{0.952} \\
    Hepatic Mineralization                                   & \textbf{0.153} & \textbf{0.902} & 0.083 & 0.866 & 0.133 & 0.872 \\
    Ingesta in the stomach                                   & \textbf{0.228} & 0.860 & 0.216 & 0.855 & 0.223 & \textbf{0.861} \\
    Irregular or granular material   in the small intestines & \textbf{0.069} & 0.844 & 0.057 & 0.838 & 0.064 & \textbf{0.857} \\
    Irregular small intestinal gas   patterns                & 0.022 & 0.852 & 0.022 & \textbf{0.853} & \textbf{0.025} & 0.839 \\
    Large Kidney                                             & 0.230 & 0.896 & 0.158 & 0.896 & \textbf{0.268} & \textbf{0.898} \\
    Limb Fracture                                            & \textbf{0.347} & \textbf{0.948} & 0.216 & 0.931 & 0.234 & 0.942 \\
    Luxation                                                 & \textbf{0.188} & \textbf{0.916} & 0.142 & 0.901 & 0.139 & 0.897 \\
    Mediastinal Mass Effect                                  & \textbf{0.402} & \textbf{0.956} & 0.371 & 0.948 & 0.393 & 0.952 \\
    Mediastinal Widening                                     & \textbf{0.521} & \textbf{0.978} & 0.484 & 0.968 & 0.521 & 0.971 \\
    Megacolon                                                & 0.010 & 0.788 & 0.015 & 0.791 & \textbf{0.027} & \textbf{0.807} \\
    Mid Abdominal Mass                                       & \textbf{0.333} & \textbf{0.954} & 0.280 & 0.940 & 0.320 & 0.930 \\
    Misshapen Kidney(s)                                      & \textbf{0.120} & \textbf{0.946} & 0.099 & 0.944 & 0.093 & 0.938 \\
    Pneumothorax                                             & \textbf{0.445} & \textbf{0.955} & 0.364 & 0.941 & 0.427 & 0.942 \\
    Prostatic Enlargement                                    & \textbf{0.358} & \textbf{0.964} & 0.277 & 0.948 & 0.326 & 0.944 \\
    Pulmonary Alveolar                                       & 0.787 & 0.975 & 0.788 & \textbf{0.977} & \textbf{0.796} & 0.976 \\
    Pulmonary Interstitial -   Nodule(s) (Under 1 cm)        & 0.335 & 0.915 & 0.320 & 0.911 & \textbf{0.361} & \textbf{0.919} \\
    Pulmonary Mass (Over 1   cm)                             & 0.392 & 0.931 & 0.349 & 0.917 & \textbf{0.409} & \textbf{0.959} \\
    Pulmonary Vascular                                       & 0.531 & \textbf{0.959} & 0.516 & 0.957 & \textbf{0.532} & 0.924 \\
    Pyloric outflow   obstruction                            & 0.039 & \textbf{0.905} & \textbf{0.054} & 0.865 & 0.044 & 0.899 \\
    Renal Mineralization                                     & \textbf{0.161} & \textbf{0.906} & 0.148 & 0.899 & 0.144 & 0.897 \\
    Rib Fracture                                             & \textbf{0.095} & \textbf{0.851} & 0.031 & 0.813 & 0.044 & 0.801 \\
    Sign(s) of IVDD                                          & \textbf{0.622} & \textbf{0.936} & 0.541 & 0.912 & 0.563 & 0.919 \\
    Sign(s) of Pleural   Effusion                            & \textbf{0.769} & \textbf{0.986} & 0.738 & 0.981 & 0.752 & 0.983 \\ 
    Small Intestinal   Obstruction                           & \textbf{0.440} & \textbf{0.960} & 0.383 & 0.944 & 0.433 & 0.954 \\
    Small Intestinal   Plication                             & 0.047 & \textbf{0.933} & 0.036 & 0.908 & \textbf{0.065} & 0.731 \\ 
    Small Kidney                                             & \textbf{0.429} & \textbf{0.963} & 0.400 & 0.961 & 0.402 & 0.960 \\
    Small Liver                                              & \textbf{0.016} & 0.773 & 0.022 & 0.744 & 0.015 & \textbf{0.896} \\
    Splenomegaly                                             & \textbf{0.450} & \textbf{0.955} & 0.403 & 0.944 & 0.393 & 0.948 \\
    Sternal Lymph Node   Enlargement                         & 0.213 & 0.936 & 0.189 & 0.936 & \textbf{0.229} & \textbf{0.944} \\
    Stifle Effusion                                          & \textbf{0.785} & \textbf{0.989} & 0.753 & 0.989 & 0.771 & 0.989 \\
    Subcutaneous Mass                                        & \textbf{0.538} & \textbf{0.949} & 0.476 & 0.931 & 0.509 & 0.889 \\
    Subcutaneous Nodule                                      & 0.076 & 0.907 & 0.061 & 0.883 & \textbf{0.082} & \textbf{0.924} \\
    Urinary Bladder   Calculus/Calculi                       & \textbf{0.339} & \textbf{0.916} & 0.329 & 0.916 & 0.317 & 0.907 \\ 
    Uterine Enlargement                                      & \textbf{0.590} & \textbf{0.976} & 0.415 & 0.933 & 0.535 & 0.949\\
    \hline
    \end{tabular}
}
\end{table}
\FloatBarrier

\end{document}